\documentclass[letterpaper, 10 pt, conference]{ieeeconf}  

\IEEEoverridecommandlockouts                              

\usepackage{graphicx}
\usepackage{hyperref}
\usepackage{xcolor}
\usepackage{subcaption}
\usepackage{cite}
\usepackage{hyperref}

\title{Show Why the Answer is Correct! Towards Explainable AI \\using Compositional Temporal Attention}

\author{Nihar Bendre, Kevin Desai and Peyman Najafirad
\thanks{Nihar Bendre, Kevin Desai and Peyman Najafirad are with AILA (Artificial Intelligent Laboratory and Autonomy),         University of Texas at San Antonio, San Antonio, Texas. USA
        {\tt\small \{nihar.bendre, kevin.desai, peyman.najafirad\}@utsa.edu}}%
}

\begin{document}

\maketitle
\thispagestyle{empty}
\pagestyle{empty}

\begin{abstract}
Visual Question Answering (VQA) models have achieved significant success in recent times. 
Despite the success of VQA models, they are mostly black-box models providing no reasoning about the predicted answer, thus raising questions for their applicability in safety-critical such as autonomous systems and cyber-security.
Current state of the art fail to better complex questions and thus are unable to exploit compositionality.
To minimize the black-box effect of these models and also to make them better exploit compositionality, we propose a Dynamic Neural Network (DMN), which can understand a particular question and then dynamically assemble various relatively shallow deep learning modules from a pool of modules to form a network. 
We incorporate compositional temporal attention to these deep learning based modules to increase compositionality exploitation.
This results in achieving better understanding of complex questions and also provides reasoning as to why the module predicts a particular answer.
Experimental analysis on the two benchmark datasets, VQA2.0 and CLEVR, depicts that our model outperforms the previous approaches for Visual Question Answering task as well as provides better reasoning, thus making it reliable for mission critical applications like safety and security.

\end{abstract}

\section{Introduction}

Visual Question Answering (VQA) \cite{zhou2020unified, kafle2017analysis} is a domain under the Computer Vision-Artificial Intelligence (AI) domain which is gathering a lot of attention from the research community.
The end-goal is to make more human-like systems that can conduct conversations and can hold knowledge of previously asked questions and their respective answers \cite{antol2015vqa, chen2020counterfactual}. 
VQA combines image-text analysis techniques where the model is tasked with predicting an answer to a textual question based on a particular image \cite{kazemi2017show, teney2018tips}.

\begin{figure}[t!]
\centering
\includegraphics[width=\linewidth]{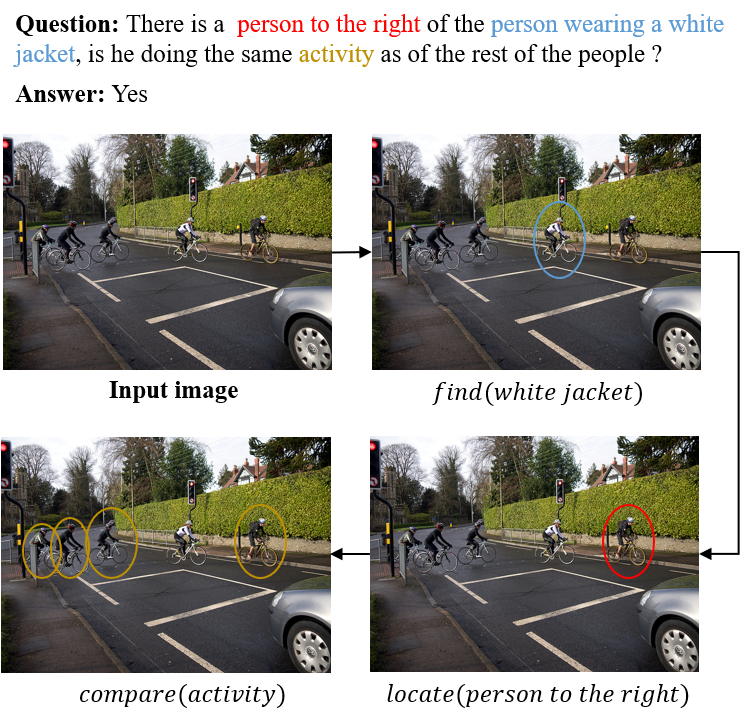}
\caption{\small{Inference example of our proposed model, Dynamic Modular Network (DMN). Our model is able to use compositional temporal attention on the question and generate a policy implementation to optimally and dynamically assemble modules to correctly predict the answer for the asked question. The highlights in the question indicate the compositional attention for policy implementation to dynamically assemble the modular network which consists of \textit{find, locate, compare} modules}} 
\label{fig:first_page}
\end{figure}

Even though deep learning based VQA models have made impressive progress towards solving the VQA challenge, they are still notorious in being a black-box system where they provide no reasoning behind the prediction \cite{andreas2016neural, gokhale2020vqa}.
This black-box approach raises the issue of trust and dependency which makes it not suitable for mission critical applications such as security, surveillance, and self-driving cars \cite{ray2017art, patro2020robust}.
Traditional VQA models also fail to exploit the compositionality in an image which can help answer complex questions, e.g., "There is a person to the right of a person wearing white jacket, is he doing the same activity as of the rest of the group?" (as shown in \autoref{fig:first_page}) \cite{tang2020semantic}.
For a VQA system to thrive, it needs the ability to exploit the compositionality to its maximum.
Recent state-of-the-art VQA models have shown their ability to exploit compositionality and also their ability to provide reasoning behind their predictions \cite{hu2017learning, silva2021adaptive, das2020opportunities}. 
This ability makes the model answer questions with some level of abstraction or questions which are not straight forward. 
Still, the compositionality exploitation is not to its maximum capacity due to the rigid structure of VQA networks. 

To some extent, this rigidity is addressed by the use of modular architecture proposed as Neural Module Network (NMN) \cite{hu2017learning, andreas2016neural}.
NMN's approach is based on using different modules dedicated for various sub-tasks and then combining them towards solving the textual question, where they use a Reinforcement Learning (RL) based policy-gradient approach to determine the layout of the modular network.
NMNs also use an external traditional linguistic parser to process the input question which it has its own limitations, e.g., their performance drops when working on vision-text tasks like VQA \cite{johnson2017clevr, bendre2020learning, xi2020visual}.
At the same time, the existing modular approach is limited to a fixed subject-object relationship in images for which they do not need a parser.

To further reduce the NMNs rigidity and also to better understand the image-text relationship, we propose a Dynamic Modular Network (DMN) which is capable of dynamically assembling modules based on the natural language based textual input question.
The key difference between the NMN approach and ours is the designing of the policy  implementation module.
Our approach includes a Recurrent Neural Network (RNN) based Long-Short Term Memory (LSTM) coupled with attention mechanism to parse the question to determine the modular network layout.
Also, instead of using a traditional linguistic parser to process the input question, where we have a LSTM network to determine the optimal policy for assembling the modules. 
These assembled modules are then applied to the input images in order to predict the answer to the particular question asked. 
Our model is able to successfully understand the textual question into various linguistic sub-structures.
Based on these sub-structures, our approach is able to dynamically compose a modular network from a pool of different compact modules dedicated for a specific task. 
The different types of modules we use are: \textit{compare, find, less than/greater than, relocate, describe, and, or, count, filter}.

\autoref{fig:first_page} illustrates an example of an image and the corresponding question to which our model is able to reason and correctly predict the answer.
The attention mechanism coupled with the different task dependent modules is able to provide adequate reasoning towards the prediction of the answer.
This overcomes the issue of trust and dependency in different applications. 
Also, the prediction accuracy is not affected by the complexity of the question.
Our proposed model is able to overcome these shortcomings and experimental analysis on two benchmark datasets (CLEVR and VQA2.0) shows that our model outperforms all the existing techniques in the safety and security domain. 


Our Contributions to this paper are as follows:
\begin{itemize}
    \item We propose a modular VQA which can answer questions related to security and safety applications.
    \item Our proposed model is able to better exploit the compositionality in images resulting in better prediction of answers for the asked questions.
\end{itemize}

\section{Related Work}

In the related work section, we focus on two aspects: visual question answering and; modular networks. 

\subsection{Visual Question Answering}

The end result of VQA \cite{malinowski2014multi, lobry2020rsvqa} is to predict answers to an natural language based textual input question taking into consideration the compositionality in the joint image-text relationship.
The performance of these models is measured against the capacity of these networks to correctly predict the answer with proper reasoning thus eliminating the traditional black-box approach.
In recent times, there has been a considerable increase in the number of datasets available to train models for VQA \cite{antol2015vqa} like the DAQUAR dataset \cite{malinowski2014multi}, COCOQA dataset \cite{yu2015visual, bendre2020human}, CLEVR dataset \cite{johnson2017clevr}, VQA and VQA2.0 datasets \cite{VQA, goyal2017making}, etc. 
The DAQUAR dataset has relatively few samples of indoor scenes thus restricting its usage towards a generic VQA model. 
The COCOQA dataset contains automatically generated question-answer pairs from images along their respective descriptions from the MSCOCO dataset. 
The CLEVR dataset contains images of objects of different shape, size, color and occlusion. 
Compared to other datasets CLEVR dataset focuses on questions which tend exploit the compositionality in images and also the reasoning ability. 
The VQA and VQA2.0 datasets contain a relatively large number image-question pairs which are crowd sourced and have a larger distribution when it comes to variety in images.
CLEVR, VQA and VQA2.0 datasets present the greatest challenges and reasoning ability for the standard VQA approaches.
A recent work by Goyal et. al \cite{goyal2017making} shows that there is a performance improvement in VQA just by memorizing the statistics related to question-answer pairs, but the downside to this approach is the limited reasoning ability due to the memorization.
Jabri et. al \cite{jabri2016revisiting} proposed a bag-of-words approach towards textual representations, however their approach yielded comparatively poor results for more sophisticated questions due to the fact that their approach had a limited exploitation of compositionality.

\subsection{Modular Network}

Neural Modular Networks (NMNs) \cite{andreas2016learning} architecture is the inspiration behind our approach, which consists of shallow task-specific recursive neural networks modules which can be assembled together dynamically.
In a generic NMN model, the model is connected with a layout policy which provides an architectural template for dynamically structuring the modules together for computation.
The training for these modules can be performed jointly for various scenarios to exploit the compositionality.
Current work on NMNs revolve around the processing of the textual data by performing linguistic analysis to generate the layout policy \cite{andreas2016neural}.
Primitive work in NMNs was based on a fixed rule-based approach for linguistic analysis to generate the layout policy.
Advancements to these works involved the use of `Dynamic Neural Module Networks' (D-NMNs) where the layout policy was generated by learning to re-rank and rearrange the neural modules based on their priority. 
The work done by Hu et. al in \cite{hu2017learning} builds on top of the D-NMNs where their approach learns to optimize the stack of neural modules instead of just re-arranging, thus eliminating the need of an external parser during testing. 
Hu et.al in \cite{hu2017modeling} propose the use of `compositional modular network' architecture where they use a parameterized resemble approach, but the scope of their approach is limited due to the fixed generation of layout policies. 

\begin{figure*}[t!]
\centering
\includegraphics[width=0.9\linewidth]{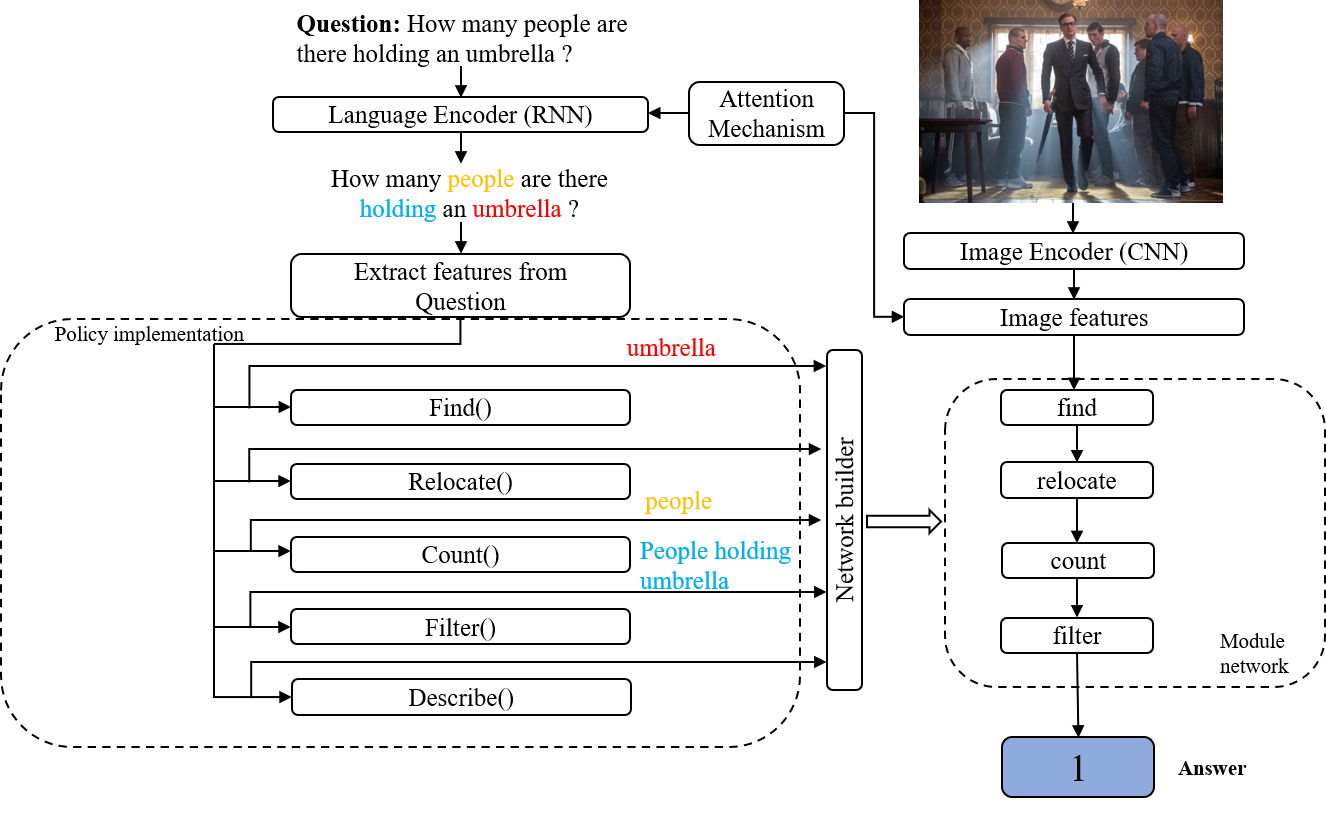}
\caption{\small{High-level illustration of the proposed modular network. In our approach, we use soft attention mechanism and self-attention mechanism along with the layout policy to dynamically assemble the structure of the modular network which can better exploit compositionality and also provide reasoning behind the predicted answer.}} 
\label{fig:overall}
\end{figure*}

Our work is built on top of the work done by \cite{hu2017learning}, where we use soft attention mechanism and self-attention mechanism along with a LSTM network to determine the layout policy to design the structure of the modular network which is fine-tuned towards the application of security and safety.  

\section{Methodology}

\autoref{fig:overall} illustrates the high-level architecture of the proposed modular network which is capable of dynamically assembling various shallow deep learning based modules based on the layout policy.
The layout policy is not dependent on a traditional language parser. Instead, it is based on a model which can predict the distribution over a much larger layout space for all possible layout options. 
The input question is encoded to its deep representation using a language encoder, which is a LSTM model.
The attention mechanism extracts a series of attentive actions from the representation, which along with the policy implementation output is passed on to the network builder block.
The policy implementation block uses this deep representation to predict the structure of the modules. 
Based on these two inputs, the network builder block dynamically assembles the fitting structure of the modules to form a neural network.
The input image is forward-passed through this dynamically assembled network to correctly predict the answer for the asked textual question.

\subsection{Attention Mechanism}
\label{sec:att}
Our proposed model consists of small individual deep learning modules which are dynamically assembled using the policy implementation based on the input question asked.
To help better exploit the compositionality, we propose to use attention mechanism over the input question.
Attention mechanism, like the name suggests, is used to highlight important words in the question.
For example, in reference to the question asked in \autoref{fig:overall}, it would be an advantage if attention is brought to words such as \textit{umbrella} and \textit{people}.
Attention mechanism takes zero or more inputs ($x_0, x_1, x_2, ... x_n$) in the form of tensors ($t_1, t_2, t_3, ...t_m$) to perform a distinct computation to output a resultant tensor ($y)$.
The overall functioning of attention mechanism can be given by:

\begin{equation}
\label{equ:l_am}
y = f(t_1, t_2, t_3, ..., t_n|x_{img}, v_{txt}, \theta)
\end{equation}

where, $x_{img}$is the image feature vector, $v_{txt}$ is the variable length vector vector of the input question, and $\theta$ is parameters of the model. 

In our approach, the tensors ($t_1, t_2, t_3, ...t_m$) are the attention maps obtained over the image features generated from the convolution operation. 
The resultant tensor ($y$) is either an attention map or the predicted answer, depending on the type of deep learning module used for a particular task. 

\autoref{tab:modules} indicates the different types of modules used, the attention inputs to them, their output (attention map or predicted answer), and the features against which the attention mechanism is applied on. 
The \textit{find} is applicable in scenarios where the need is to find/localize any object or human. 
The \textit{compare, greater than, less than, equal to} modules are applicable in scenarios where the need is to perform a comparison between two entities. 
The \textit{describe} module is used in scenarios where the need is to extract additional complex details.
The modules \textit{count} and \textit{exists} are used when we need to determine if a particular entity exists and if yes,  then the count module performs a quantitative analysis.
The \textit{relocate} module is used to exploit the compositionality by transforming the input attention feature maps into a modified feature map.
The \textit{filter, or, and} modules are helpful to clarify the policy implementation.

\begin{table}[t!]
\centering
\begin{tabular}{l|c|c|c}  \hline
\textbf{Module}       & \textbf{Attention}  & \textbf{Output}           & \textbf{Extracted Features} \\ \hline
find         & -          & Attention        & $x_{img}$, $v_{txt}$     \\
compare      & $t_1$, $t_2$ & Prediction & $x_{img}$, $v_{txt}$    \\
describe     & $t$          & Prediction & $x_{img}$, $v_{txt}$   \\
exist        & $t$          & Prediction & -                \\
equal to     & $t$          & Prediction & -               \\
and          & $t_1$, $t_2$ & Attention        & -              \\
filter       & $t $         & Attention        & $x_{img}$, $v_{txt}$   \\
relocate     & $t $         & Attention        & $x_{img}$, $v_{txt}$   \\
or           & $t_1$, $t_2$ & Attention        & -               \\
greater than & $t_1$, $t_2$ & Prediction & -               \\
less than    & $t_1$, $t_2$ & Prediction & -                \\
is present   & $t$          & Prediction & -                \\ 
count        & $t$          & Prediction & -  \\ \hline           
\end{tabular}
\caption{\small{Tables illustrating the different modules that are available for dynamically assembling to predict the answer for the particular question asked.}}
\label{tab:modules}
\end{table}

\subsection{Policy Implementation}
\label{sec:pol_imp}
The overall objective of the policy implementation block is to predict a custom fit structure of modules from the available pool of modules for the question asked. 
For example, to a question ($q =$) `What animal is the person playing with ?', the policy implementation block will output an probability distribution $P(p_i|q)$ that can best answer the question.
Here, $p_i$ corresponds to the optimal policy implementation based on which a neural network is dynamically assembled from the pool of available modules.
Previous approaches \cite{andreas2016learning, andreas2016neural} use an external parser to determine the policy layout, thus making it restricted to the scope of the parser.
In comparison, our approach can search over a broader policy distribution (almost all possible policy implementation outcomes) to come up with the most suitable policy ($p_i$).
Similar to the work done in \cite{hu2017learning}, we use Reverse Polish Notation \cite{burks1954analysis} to represent the policy sequence in a linear format by mapping them to a syntax tree one at a time.
The motivation behind transforming the modules into a linear fashion is to look at it as a sequence-to-sequence problem starting from the input question ($q$) and down to the assembling of the modules.

Recurrent Neural Networks (RNN) have shown to achieve state-of-the-art approach in solving sequence-to-sequence problems \cite{bahdanau2014neural}.
Thus, we use a modified version of RNN called Long-Short Term Memory (LSTM) towards the training of the policy implementation block.
The language encoder block consists of a LSTM network which performs a word-wise encoding of the input question to a variable-length vector ($v$).
The length of the vector is equal to the number of words in the input question.
This vector is then decoded using a LSTM which is composed of similar structure as that of the encoding LSTM, with the difference being in the use of different parameters.
While encoding, we a apply soft attention mechanism (Section \ref{sec:att}) over the vector and the input question and during the decoding process we use the attention weights generated by the attention mechanism to predict the structure of the optimal policy implementation.

\begin{figure*}[t!]
\centering
\includegraphics[width=0.85\linewidth]{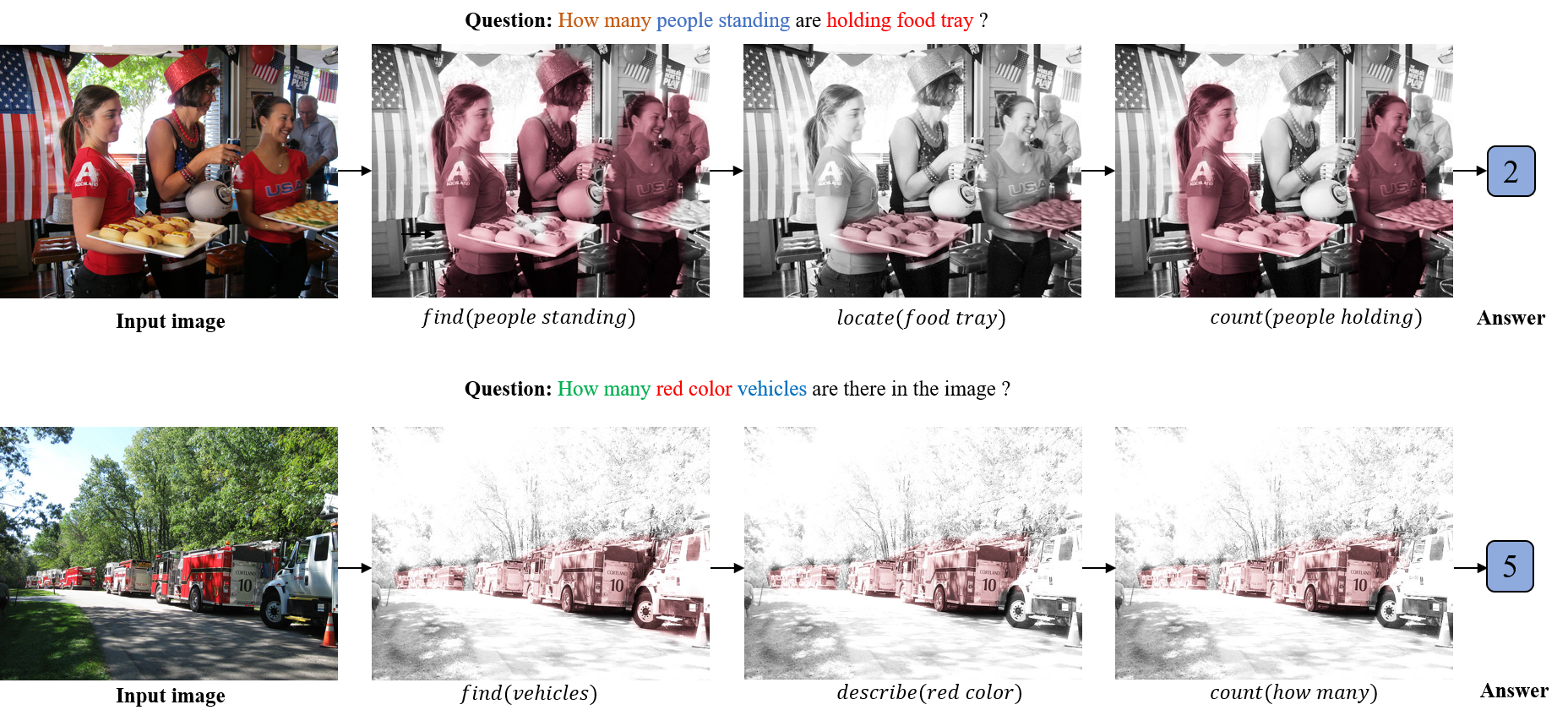}
\caption{\small{Examples of prediction performed using our model. We can see that the module dynamically assembles the required modules to form the network to predict the correct answer for the question asked. We also provide the ground truth for comparison.} }
\label{fig:examples}
\end{figure*}

\subsection{Modular Network}
\label{sec:Mod_net}
The goal of the modular network block is to dynamically assemble during training depending upon the input question asked.
This is achieved by optimizing the loss during the policy implementation (Section \ref{sec:pol_imp}).
The optimal policy for assembling the modular network is determined by simultaneously optimizing (reducing the loss) the network parameters and the policy implementation. 
The overall loss function during training is given by:

\begin{equation}
\label{equ:l_cm}
\mathcal{L}_{\theta} = \mathcal{E}_{P(p_i|q;\theta)}[\hat{L}(\theta, p; q, i]
\end{equation}

where, $i$ is the input image, $q$ is the question asked, $p$ is the policy implementation and $P$ is the probability distribution. During back propagation, the $\mathcal{L}_{\theta}$ is optimized by using policy gradient method commonly used in reinforcement learning and using the Monte-Carlo method the gradient is determined.

\begin{table*}[t!]
\centering
\begin{tabular}{l|c|c|c|c|c} \hline
\textbf{Model}                                                 & \textbf{Overall Acc (\%)} & \textbf{Exist Acc (\%)} & \textbf{Count Acc (\%)} & \textbf{Yes/No Acc (\%)} & \textbf{Compare Acc (\%)} \\ \hline
CNN+BoW \cite{zhou2015simple}                                               & 48.4             & 59.5           & 38.9           & 59.0            & 47.0             \\
CNN+LSTM  \cite{antol2015vqa}                                             & 52.3             & 44.1           & 45.1           & 56.0            & 38.0             \\
CNN+LSTM+MCB \cite{fukui2016multimodal}                                          & 51.4             & 64.3           & 44.2           & 58.0            & 49.0             \\
CNN+LSTM+SA \cite{yang2016stacked}                                            & 69.6             & 78.4           & 50.1           & 79.8            & 51.6             \\
Neural Module Network \cite{andreas2016neural}                                 & 73.2             & 81.2           & 55.7           & 83.5            & 55.7             \\
End-to-End Neural Module Network  \cite{hu2017learning}                     & 83.9             & 86.0           & 61.9           & 85.5            & 60.9             \\ \hline
Baseline I - DMN w/o Attention w/ LSTM           & 76.7             & 57.0           & 46.5           & 68.1            & 47.3             \\
Baseline II - DMN w/ Attention w/o LSTM  & 78.1             & 67.2           & 53.6           & 77.1            & 49.7             \\
\textbf{Dynamic Modular Network (DMN)}                           & \textbf{86.9}             & \textbf{87.4}           & \textbf{66.0 }          & \textbf{88.9 }           & \textbf{67.2}    \\ \hline        
\end{tabular}
\caption{\small{Experimental Analysis of our model, Dynamic Neural Network (DMN) on the VQA2.0 dataset and comparison of our model to the previous approaches. The lower half of the table highlights the ablation study of our model, where Baseline I is the model where we do not use attention mechanism and Baseline II is a model where we use the traditional language parser instead of our approach for policy implementation.}}
\label{tab:exp_clvr}
\end{table*}

\begin{table}[t!]
\centering
\begin{tabular}{l|c} \hline
\textbf{Model}                               & \textbf{Acc (\%)} \\ \hline
CNN+LSTM+MCB  \cite{fukui2016multimodal}                                & 64.7                   \\
D-NMN \cite{andreas2016learning}                                        & 57.9                   \\
Neural Module Network (NMN) \cite{andreas2016neural}                  & 57.3                   \\
End-to-End NMN  \cite{hu2017learning}                              & 64.2                   \\ \hline
Baseline I - DMN w/o Attention w/ LSTM                      & 51.4                   \\
Baseline II - DMN w/ Attention w/o LSTM             & 53.9                   \\
\textbf{Dynamic Modular Network (DMN)} & \textbf{66.8}      \\ \hline   
\end{tabular}
\caption{\small{Experimental Analysis of our model, Dynamic Neural Network (DMN) on the VQA2.0 dataset and comparison of our model to the previous approaches. The lower half of the table highlights the ablation study of our model, where Baseline I is the model where we do not use attention mechanism and Baseline II is a model where we use the traditional language parser instead of our approach for policy implementation.}}
\label{tab:exp_vqa}
\end{table}


\section{Experiments}

To determine our model's performance and its ability to exploit compositionality and provide reasoning we test our model on two publicly available large scale benchmark datasets: VQA2.0 and CLEVR. 

\subsection{Experimental Results}

We use Adam Optimizer \cite{kingma2014adam} and back-propagation to train our model end-to-end. The training was performed on two Nvidia Titan 1080ti GPUs. To obtain the image features from CLEVR dataset we use a (pretrained on ImageNet) VGG-19 model \cite{simonyan2014very} whereas for the VQA2.0 dataset we use ResNet-152 model \cite{he2016deep} which is also pretrained on ImageNet.


VQA2.0 dataset contains a relatively large number image-question pairs which are crowd sourced and have a larger distribution when it comes to variety in images. 
Having such a wide distribution of images-questions helps in making the model more robust especially for the safety and security applications. 
For training, we stick to the original training-testing split provided where we use 443,800 questions based on 82,800 images for training, 214,400 questions based on 40,500 questions for validation/testing.
\autoref{tab:exp_vqa} highlights the performance of our model on the VQA2.0 dataset by comparing our approach to the previously state-of-the-art techniques which use a similar concept of modular networks.


The CLEVR dataset contains images of objects of different shape, size, color and occlusion. 
CLEVR dataset consists of 853,400 questions based on 100,000 images.
Compared to other datasets CLEVR dataset focuses on questions which tend exploit the compositionality in images and also the reasoning ability. 
for example, `There is a small gray block; are there any spheres to the left of it' \cite{johnson2017clevr}. 
The questions tend to have a longer length as well which makes its understanding by the LSTM more challenging \cite{vaswani2017attention} and requires complex understanding for inference. 
\autoref{tab:exp_clvr} highlights the evaluation of our model on the CLEVR dataset by comparing our approach to the previously state-of-the-art techniques which use a similar concept of modular networks.
We also evaluated the model's ability to solve a particular situation of question, like \textit{exist, count, yes/no, compare}.

\subsection{Ablation Study}

\autoref{tab:exp_clvr} and \autoref{tab:exp_vqa} highlight the ablation study of our model on the benchmark datasets: CLEVR and VQA2.0 respectively.
We used the same baseline models for testing the model's performance on both the datasets

\textbf{Baseline I:} We used the same model with the LSTM as the parser but without the attention mechanism. 
The results without the attention mechanism, neither soft nor self-attention as (discussed in \autoref{sec:att}) show a significant reduction in the accuracy of the model for both the datasets.

\textbf{Baseline II:} We replace the LSTM parser with the traditional text parser and apply apply attention mechanism to generate the optimal policy implementation instead of the Recurrent Neural Network based LSTM approach as discussed in \autoref{sec:pol_imp}.

Performing a direct comparison between the results from Baseline I and Baseline II, we can assertively say that the attention mechanism along LSTM network for optimal selection of policy implementation layout has a direct effect on the exploitation of compositionality thus affecting the overall prediction. 
Attention mechanism also provides helps in better understanding of the asked question and provide reasoning behind the predicted answer to the asked question.

\subsection{Discussion and Future Work}

\begin{figure*}[t!]
\centering
\includegraphics[width=0.85\linewidth]{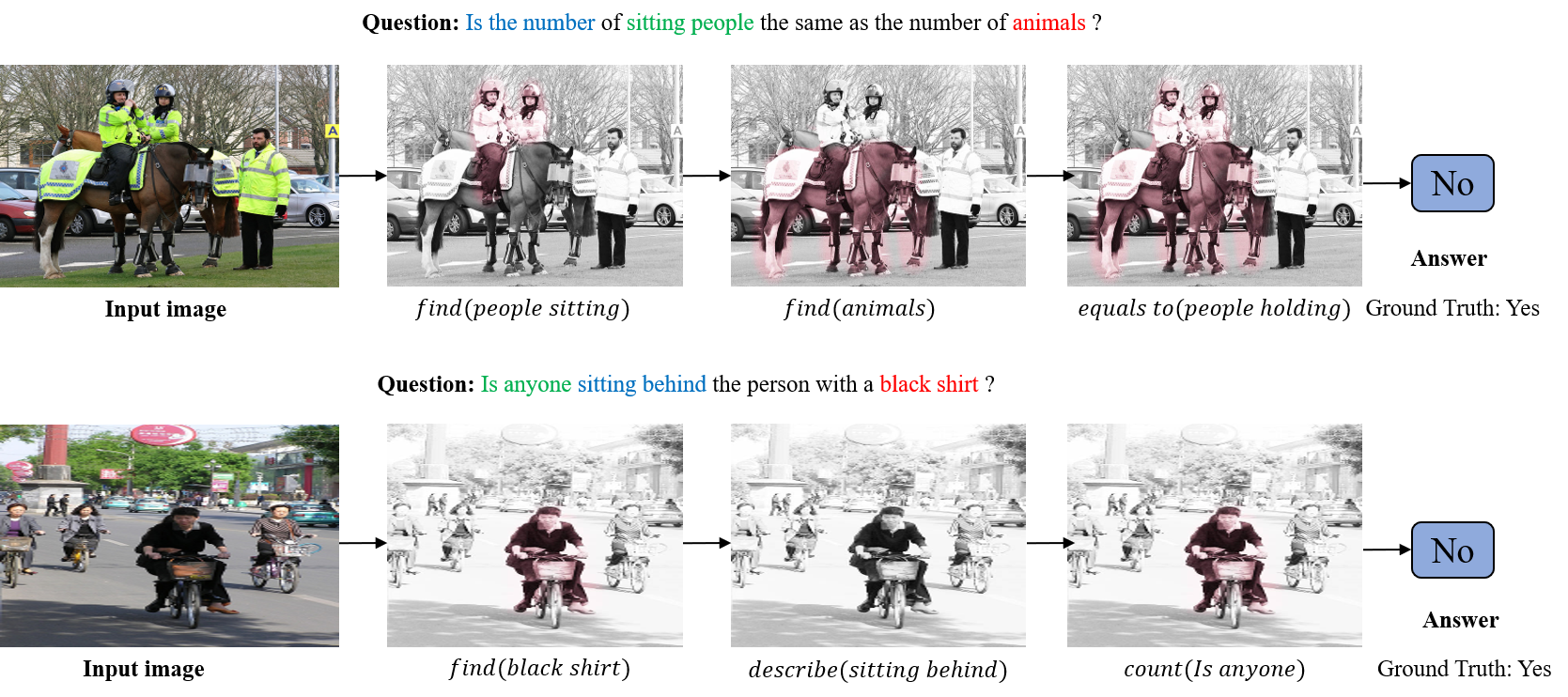}
\caption{\small{We highlight few cases where our model fails to predict the correct answer. The reason behind our model's failure is heavy occlusion.}}
\label{fig:discussion}
\end{figure*}

\autoref{fig:discussion} highlights few of the cases where our proposed model failed to predict the correct answer.
The image of the left shows a bunch of people cycling on a road. 
Our naked eye can tell that behind the lead cyclist there is a person sitting as well.
And when asked the question `Is anyone sitting behind the person with a black shirt ?', the model  predicts wrongly. 
The reason behind the wrong prediction is the occlusion of the person sitting behind which is also the reason why there is no attention shown for the function of \textit{describe(sitting behind)}.
Due to the heavy occlusion, our model is not able to distinguish it as two different people and instead considers them as one. 
The image on the right depicts that two people are sitting on top of two horses and another person is standing next to them.
In this case as well, there is a considerable amount of occlusion between the two horses and therefore our model cannot distinguish them resulting in a wrong prediction to the question, `Is the number of sitting people the same as the number of animals ?'
In reference to \autoref{fig:examples}, where the image consists of a bunch of fire trucks parked alongside a road and the question asked was `How many red color vehicles are there in the image ?', to which our model was able to correctly predict the answer.
This implies that, irrespective of the size of vehicle going from big to small, our model was able to correctly predict, thus confirming that our model was able to exploit the compositionality in the image.

As part of future work, we want to extend our model, where the model is able to predict an entire sentence as an answer to the asked question and thus a step closer to solving the conversational question answering challenge.
Also we would like to extend our model to work on videos where it can exploit the temporal information and correctly predict the answer.

\section{Conclusion}

In this research work, we introduce a Dynamic Modular Network (DMN) towards solving the challenging task of Visual Question Answering (VQA).
Our model is able to process an image and a textual question related to the image, and predict the answer to the question. 
We propose the use of a LSTM network coupled with attention mechanism to better understand the textual question.
Based on this, the policy implementation block is able to dynamically assemble the corresponding modules.
We use attention mechanisms (soft and self-attention) to better exploit the compositionality and for the better understanding of the questions.
As the modules are dynamically assembled and are structurally different for different questions, the assembling of these such modular network provide a reasoning towards the prediction of the answer for any given set of an image and its related question.
Experimental Analysis on the two benchmark datasets: VQA2.0 and CLEVR highlight that our model outperforms the previous approaches for the challenging task of Visual Question Answering (VQA).


\bibliographystyle{IEEEtran}
\bibliography{root}

\begin{thebibliography}{10}
\providecommand{\url}[1]{#1}
\csname url@samestyle\endcsname
\providecommand{\newblock}{\relax}
\providecommand{\bibinfo}[2]{#2}
\providecommand{\BIBentrySTDinterwordspacing}{\spaceskip=0pt\relax}
\providecommand{\BIBentryALTinterwordstretchfactor}{4}
\providecommand{\BIBentryALTinterwordspacing}{\spaceskip=\fontdimen2\font plus
\BIBentryALTinterwordstretchfactor\fontdimen3\font minus
  \fontdimen4\font\relax}
\providecommand{\BIBforeignlanguage}[2]{{%
\expandafter\ifx\csname l@#1\endcsname\relax
\typeout{** WARNING: IEEEtran.bst: No hyphenation pattern has been}%
\typeout{** loaded for the language `#1'. Using the pattern for}%
\typeout{** the default language instead.}%
\else
\language=\csname l@#1\endcsname
\fi
#2}}
\providecommand{\BIBdecl}{\relax}
\BIBdecl

\bibitem{zhou2020unified}
L.~Zhou, H.~Palangi, L.~Zhang, H.~Hu, J.~Corso, and J.~Gao, ``Unified
  vision-language pre-training for image captioning and vqa,'' in
  \emph{Proceedings of the AAAI Conference on Artificial Intelligence},
  vol.~34, no.~07, 2020, pp. 13\,041--13\,049.

\bibitem{kafle2017analysis}
K.~Kafle and C.~Kanan, ``An analysis of visual question answering algorithms,''
  in \emph{Proceedings of the IEEE International Conference on Computer
  Vision}, 2017, pp. 1965--1973.

\bibitem{antol2015vqa}
S.~Antol, A.~Agrawal, J.~Lu, M.~Mitchell, D.~Batra, C.~L. Zitnick, and
  D.~Parikh, ``Vqa: Visual question answering,'' in \emph{Proceedings of the
  IEEE international conference on computer vision}, 2015, pp. 2425--2433.

\bibitem{chen2020counterfactual}
L.~Chen, X.~Yan, J.~Xiao, H.~Zhang, S.~Pu, and Y.~Zhuang, ``Counterfactual
  samples synthesizing for robust visual question answering,'' in
  \emph{Proceedings of the IEEE/CVF Conference on Computer Vision and Pattern
  Recognition}, 2020, pp. 10\,800--10\,809.

\bibitem{kazemi2017show}
V.~Kazemi and A.~Elqursh, ``Show, ask, attend, and answer: A strong baseline
  for visual question answering,'' \emph{arXiv preprint arXiv:1704.03162},
  2017.

\bibitem{teney2018tips}
D.~Teney, P.~Anderson, X.~He, and A.~Van Den~Hengel, ``Tips and tricks for
  visual question answering: Learnings from the 2017 challenge,'' in
  \emph{Proceedings of the IEEE conference on computer vision and pattern
  recognition}, 2018, pp. 4223--4232.

\bibitem{andreas2016neural}
J.~Andreas, M.~Rohrbach, T.~Darrell, and D.~Klein, ``Neural module networks,''
  in \emph{Proceedings of the IEEE conference on computer vision and pattern
  recognition}, 2016, pp. 39--48.

\bibitem{gokhale2020vqa}
T.~Gokhale, P.~Banerjee, C.~Baral, and Y.~Yang, ``Vqa-lol: Visual question
  answering under the lens of logic,'' in \emph{European Conference on Computer
  Vision}.\hskip 1em plus 0.5em minus 0.4em\relax Springer, 2020, pp. 379--396.

\bibitem{ray2017art}
A.~Ray, ``The art of deep connection-towards natural and pragmatic
  conversational agent interactions,'' Ph.D. dissertation, Virginia Tech, 2017.

\bibitem{patro2020robust}
B.~Patro, S.~Patel, and V.~Namboodiri, ``Robust explanations for visual
  question answering,'' in \emph{Proceedings of the IEEE/CVF Winter Conference
  on Applications of Computer Vision}, 2020, pp. 1577--1586.

\bibitem{tang2020semantic}
R.~Tang, C.~Ma, W.~E. Zhang, Q.~Wu, and X.~Yang, ``Semantic equivalent
  adversarial data augmentation for visual question answering,'' in
  \emph{European Conference on Computer Vision}.\hskip 1em plus 0.5em minus
  0.4em\relax Springer, 2020, pp. 437--453.

\bibitem{hu2017learning}
R.~Hu, J.~Andreas, M.~Rohrbach, T.~Darrell, and K.~Saenko, ``Learning to
  reason: End-to-end module networks for visual question answering,'' in
  \emph{Proceedings of the IEEE International Conference on Computer Vision},
  2017, pp. 804--813.

\bibitem{silva2021adaptive}
S.~H. Silva, A.~Das, I.~Scarff, and P.~Najafirad, ``Adaptive clustering of
  robust semantic representations for adversarial image purification,''
  \emph{arXiv preprint arXiv:2104.02155}, 2021.

\bibitem{das2020opportunities}
A.~Das and P.~Rad, ``Opportunities and challenges in explainable artificial
  intelligence (xai): A survey,'' \emph{arXiv preprint arXiv:2006.11371}, 2020.

\bibitem{johnson2017clevr}
J.~Johnson, B.~Hariharan, L.~Van Der~Maaten, L.~Fei-Fei, C.~Lawrence~Zitnick,
  and R.~Girshick, ``Clevr: A diagnostic dataset for compositional language and
  elementary visual reasoning,'' in \emph{Proceedings of the IEEE Conference on
  Computer Vision and Pattern Recognition}, 2017, pp. 2901--2910.

\bibitem{bendre2020learning}
N.~Bendre, H.~T. Mar{\'\i}n, and P.~Najafirad, ``Learning from few samples: A
  survey,'' \emph{arXiv preprint arXiv:2007.15484}, 2020.

\bibitem{xi2020visual}
Y.~Xi, Y.~Zhang, S.~Ding, and S.~Wan, ``Visual question answering model based
  on visual relationship detection,'' \emph{Signal Processing: Image
  Communication}, vol.~80, p. 115648, 2020.

\bibitem{malinowski2014multi}
M.~Malinowski and M.~Fritz, ``A multi-world approach to question answering
  about real-world scenes based on uncertain input,'' \emph{arXiv preprint
  arXiv:1410.0210}, 2014.

\bibitem{lobry2020rsvqa}
S.~Lobry, D.~Marcos, J.~Murray, and D.~Tuia, ``Rsvqa: Visual question answering
  for remote sensing data,'' \emph{IEEE Transactions on Geoscience and Remote
  Sensing}, vol.~58, no.~12, pp. 8555--8566, 2020.

\bibitem{yu2015visual}
L.~Yu, E.~Park, A.~C. Berg, and T.~L. Berg, ``Visual madlibs: Fill in the blank
  image generation and question answering,'' \emph{arXiv preprint
  arXiv:1506.00278}, 2015.

\bibitem{bendre2020human}
N.~Bendre, N.~Ebadi, J.~J. Prevost, and P.~Najafirad, ``Human action
  performance using deep neuro-fuzzy recurrent attention model,'' \emph{IEEE
  Access}, vol.~8, pp. 57\,749--57\,761, 2020.

\bibitem{VQA}
S.~Antol, A.~Agrawal, J.~Lu, M.~Mitchell, D.~Batra, C.~L. Zitnick, and
  D.~Parikh, ``{VQA}: {V}isual {Q}uestion {A}nswering,'' in \emph{International
  Conference on Computer Vision (ICCV)}, 2015.

\bibitem{goyal2017making}
Y.~Goyal, T.~Khot, D.~Summers-Stay, D.~Batra, and D.~Parikh, ``Making the v in
  vqa matter: Elevating the role of image understanding in visual question
  answering,'' in \emph{Proceedings of the IEEE Conference on Computer Vision
  and Pattern Recognition}, 2017, pp. 6904--6913.

\bibitem{jabri2016revisiting}
A.~Jabri, A.~Joulin, and L.~Van Der~Maaten, ``Revisiting visual question
  answering baselines,'' in \emph{European conference on computer
  vision}.\hskip 1em plus 0.5em minus 0.4em\relax Springer, 2016, pp. 727--739.

\bibitem{andreas2016learning}
J.~Andreas, M.~Rohrbach, T.~Darrell, and D.~Klein, ``Learning to compose neural
  networks for question answering,'' \emph{arXiv preprint arXiv:1601.01705},
  2016.

\bibitem{hu2017modeling}
R.~Hu, M.~Rohrbach, J.~Andreas, T.~Darrell, and K.~Saenko, ``Modeling
  relationships in referential expressions with compositional modular
  networks,'' in \emph{Proceedings of the IEEE Conference on Computer Vision
  and Pattern Recognition}, 2017, pp. 1115--1124.

\bibitem{burks1954analysis}
A.~W. Burks, D.~W. Warren, and J.~B. Wright, ``An analysis of a logical machine
  using parenthesis-free notation,'' \emph{Mathematical tables and other aids
  to computation}, vol.~8, no.~46, pp. 53--57, 1954.

\bibitem{bahdanau2014neural}
D.~Bahdanau, K.~Cho, and Y.~Bengio, ``Neural machine translation by jointly
  learning to align and translate,'' \emph{arXiv preprint arXiv:1409.0473},
  2014.

\bibitem{zhou2015simple}
B.~Zhou, Y.~Tian, S.~Sukhbaatar, A.~Szlam, and R.~Fergus, ``Simple baseline for
  visual question answering,'' \emph{arXiv preprint arXiv:1512.02167}, 2015.

\bibitem{fukui2016multimodal}
A.~Fukui, D.~H. Park, D.~Yang, A.~Rohrbach, T.~Darrell, and M.~Rohrbach,
  ``Multimodal compact bilinear pooling for visual question answering and
  visual grounding,'' \emph{arXiv preprint arXiv:1606.01847}, 2016.

\bibitem{yang2016stacked}
Z.~Yang, X.~He, J.~Gao, L.~Deng, and A.~Smola, ``Stacked attention networks for
  image question answering,'' in \emph{Proceedings of the IEEE conference on
  computer vision and pattern recognition}, 2016, pp. 21--29.

\bibitem{kingma2014adam}
D.~P. Kingma and J.~Ba, ``Adam: A method for stochastic optimization,''
  \emph{arXiv preprint arXiv:1412.6980}, 2014.

\bibitem{simonyan2014very}
K.~Simonyan and A.~Zisserman, ``Very deep convolutional networks for
  large-scale image recognition,'' \emph{arXiv preprint arXiv:1409.1556}, 2014.

\bibitem{he2016deep}
K.~He, X.~Zhang, S.~Ren, and J.~Sun, ``Deep residual learning for image
  recognition,'' in \emph{Proceedings of the IEEE conference on computer vision
  and pattern recognition}, 2016, pp. 770--778.

\bibitem{vaswani2017attention}
A.~Vaswani, N.~Shazeer, N.~Parmar, J.~Uszkoreit, L.~Jones, A.~N. Gomez,
  L.~Kaiser, and I.~Polosukhin, ``Attention is all you need,'' \emph{arXiv
  preprint arXiv:1706.03762}, 2017.

\end{thebibliography}

\end{document}